\documentclass[pdflatex,sn-mathphys-num,oneside]{sn-jnl}

\usepackage{graphicx}%
\usepackage{multirow}%
\usepackage{amsmath,amssymb,amsfonts}%
\usepackage{amsthm}%
\usepackage{mathrsfs}%
\usepackage[title]{appendix}%
\usepackage{xcolor}%
\usepackage{textcomp}%
\usepackage{manyfoot}%
\usepackage{booktabs}%
\usepackage{algorithm}%
\usepackage{algorithmicx}%
\usepackage{algpseudocode}%
\usepackage{listings}%
\usepackage{makecell}
\usepackage{xcolor}
\usepackage{alltt}
\usepackage[most]{tcolorbox}
\tcbuselibrary{breakable,skins,listings}
\usepackage{fvextra}
\usepackage{enumitem}
\definecolor{techXML}{HTML}{1F77B4}   
\definecolor{techCOT}{HTML}{FF7F0E}   
\definecolor{techVAL}{HTML}{D62728}   
\definecolor{techLOG}{HTML}{17BECF}   
\definecolor{techDIS}{HTML}{9467BD}   
\definecolor{techPERF}{HTML}{2CA02C}  
\newcommand{\XML}[1]{\textcolor{techXML}{#1}}
\newcommand{\COT}[1]{\textcolor{techCOT}{#1}}
\newcommand{\VAL}[1]{\textcolor{techVAL}{#1}}
\newcommand{\LOGIC}[1]{\textcolor{techLOG}{#1}}
\newcommand{\DIS}[1]{\textcolor{techDIS}{#1}}
\newcommand{\PERF}[1]{\textcolor{techPERF}{#1}}

\newcommand{\LegendItem}[2]{\textcolor{#1}{\rule{1.15ex}{1.15ex}}\hspace{0.5em}#2}
\newtcolorbox{legendbox}{
  enhanced,
  colback=white,
  colframe=black!35,
  boxrule=0.6pt,
  arc=1.2mm,
  left=5pt,right=5pt,top=4pt,bottom=4pt,
}

\theoremstyle{thmstyleone}%
\theoremstyle{thmstyletwo}%

\theoremstyle{thmstylethree}%

\begin{document}

\title[Article Title]{PVminerLLM: Structured Extraction of Patient Voice from Patient-Generated Text using Large Language Models}



\author*[1]{\fnm{Samah} \sur{Fodeh}}\email{samah.fodeh@yale.edu}
\author[1]{\fnm{Linhai} \sur{Ma}}
\author[1]{\fnm{Ganesh} \sur{Puthiaraju}}
\author[1]{\fnm{Srivani} \sur{Talakokkul}}
\author[1]{\fnm{Afshan} \sur{Khan}}
\author[1]{\fnm{Ashley} \sur{Hagaman}}
\author[1]{\fnm{Sarah} \sur{Lowe}}
\author[2]{\fnm{Aimee} \sur{Roundtree}}



\affil[1]{\orgname{Yale University}, \orgaddress{\city{New Haven}, \state{CT}, \country{USA}}}

\affil[2]{\orgname{Texas State University}, \orgaddress{\city{San Marcos}, \state{TX}, \country{USA}}}





\abstract{
\textbf{Motivation:}
Patient-generated text contains critical information about patients’ lived experiences, social circumstances, and engagement in care, including factors that strongly influence adherence, care coordination, and health equity. However, these patient voice signals are rarely available in structured form, limiting their use in patient-centered outcomes research and clinical quality improvement. Reliable extraction of such information is therefore essential for understanding and addressing non-clinical drivers of health outcomes at scale.

\textbf{Results:}
We introduce PVminer, a benchmark for structured extraction of patient voice, and propose PVminerLLM, a supervised fine-tuned large language model tailored to this task. Across multiple datasets and model sizes, PVminerLLM substantially outperforms prompt-based baselines, achieving up to 83.82\% F1 for Code prediction, 80.74\% F1 for Sub-code prediction, and 87.03\% F1 for evidence Span extraction. Notably, strong performance is achieved even with smaller models, demonstrating that reliable patient voice extraction is feasible without extreme model scale. These results enable scalable analysis of social and experiential signals embedded in patient-generated text.

\textbf{Availability and Implementation:}
Code, evaluation scripts, and trained LLMs will be released publicly. Annotated datasets will be made available upon request for research use.
}

\keywords{Large Language Models, Supervised Fine-Tuning, Medical Annotation, Patient-Generated Text, Clinical NLP}



\maketitle

\section{Introduction}\label{sec1}

\begin{figure*}[!h]
  \centering
    \includegraphics[width=\textwidth]{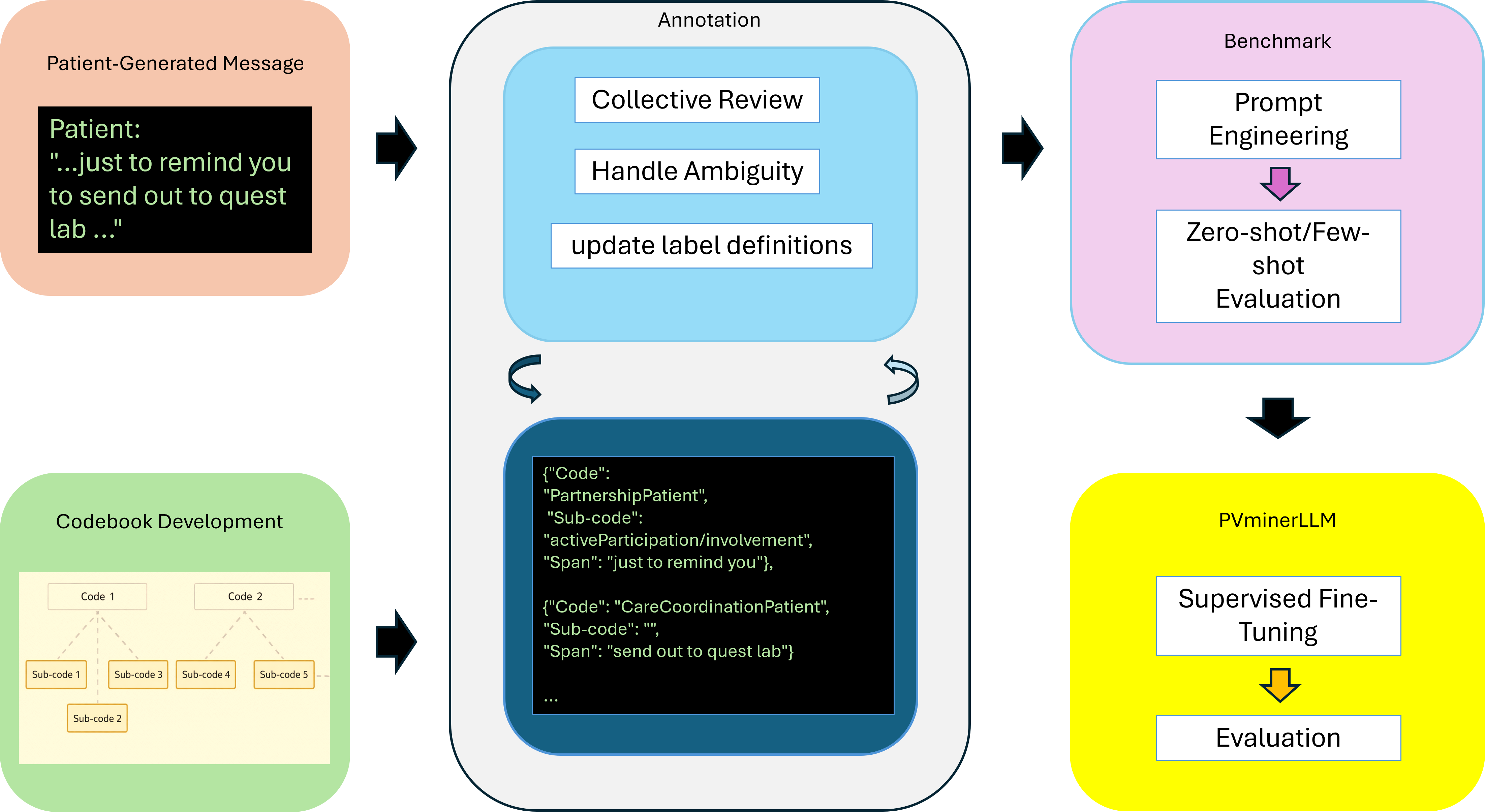}
  \caption{Overview of the PVminer Framework's pipeline: Codebook Development\(\rightarrow\) Annotation \(\rightarrow\) Prompt Engineering \(\rightarrow\) Supervised Fine-tuning}
  \label{fig:pipeline}
\end{figure*}

Patient-generated data, including secure messages, survey responses, and interview narratives, provide a rare and direct window into patients’ lived experiences outside traditional clinical encounters \cite{intro1, intro2, intro3, intro4}. Unlike structured clinical records, these texts capture how individuals articulate their needs, constraints, emotions, and expectations in their own words. Collectively, such expressions constitute the patient voice, a socially embedded signal that reflects not only clinical concerns but also broader social, environmental, and relational contexts that shape health outcomes. These contexts include social determinants of health, such as housing instability, and financial insecurity, as well as patients’ engagement, preferences, and participation in managing their care \cite{intro5, intro6, intro7, fodeh2026pvminer, fodeh2026eppcminerben, fodeh2026tabpo}.

Despite its importance, the patient voice remains largely underexplored in patient-centered outcomes research and health services research. When patient-generated text is analyzed, it is often reduced to narrow, isolated categories or surface-level entities, obscuring the complex and overlapping social realities embedded in everyday language. For example, a single patient message can simultaneously convey medical uncertainty, emotional distress, social constraints, and preferences regarding care decisions. Yet most computational approaches fail to preserve this richness, instead focusing on a limited subset of social factors \cite{intro17, intro30, intro19, intro20, intro21, intro22, intro23, intro24, intro25, intro26, intro27, intro28, intro29}. As a result, critical social signals that influence adherence, care coordination, partnership and shared decision making, and equity are systematically underrepresented in large-scale analyses \cite{intro42, intro43, raisa2023identifying}.


For conditions such as mental health and substance use disorders, where much of care occurs outside traditional clinical settings and depends heavily on sustained patient engagement, outcomes hinge not only on access to therapy or medication but also on patients’ ability to manage stigma, social stressors, and fluctuating motivation in daily life. Evidence shows that treatment adherence and continuity of care in depression, anxiety, and opioid use disorder are strongly influenced by psychosocial factors, such as housing, food, financial insecurity, stress, and perceived social support.
\cite{lopez2025quality,seaberg2025retention} Scalable methods that can reliably extract these signals from unstructured text are therefore essential for understanding treatment effectiveness and for designing responsive interventions to patients’ needs. However, manual abstraction of the patient voice from textual data and its transformation into a structured, accessible format is laborious and expensive \cite{intro11, intro12, intro13, intro14, intro15, intro16, intro17, roter1997rias}. Existing machine learning methods \cite{intro31, intro32, intro33, intro34, intro35, intro36, intro37, intro44, intro45, intro46, intro47, intro48} offer scalability but have primarily focused on clinical notes in electronic health records rather than unstructured patient-generated text, and typically target only a limited set of social domains.

In this study, we introduce the PVminer framework, which formalizes the patient voice annotation as schema-constrained structured prediction from unstructured patient-generated text. The task requires extracting hierarchical labels with the main domains including: partnership and building rapport; shared decision-making; socioemotional support; and social
determinants of health, and associated evidence Spans.
PVminer reflects intrinsic properties of patient-generated data, including highly unstructured language, a wide range of social factors, overlapping categories, severe label imbalance, and a small number of semantically critical tokens that determine annotation correctness.


To provide an initial solution using state-of-the-art scalable techniques, we first implement a dedicated prompt engineering approach and benchmark a range of instruction-tuned large language models spanning model sizes from 1.5B to 70B parameters under zero-shot and few-shot settings. Carefully engineered prompts enable these models to capture coarse semantic signals and yield measurable performance improvements. However, they frequently produce poorly structured, verbose, or truncated outputs. This behavior leads to substantial precision–recall gaps, indicating that prompting alone is insufficient for reliable patient voice extraction and motivating task-specific model adaptation.


To address these limitations, we introduce PVminerLLM, a suite of supervised fine-tuned language models specialized for the PVminer task. By adapting instruction-tuned models to the PVminer schema, PVminerLLM enforces structured, schema-valid outputs and achieves strong performance across hierarchical labels and evidence Spans. Our results demonstrate that supervised fine-tuning offers a scalable and effective approach for high-fidelity extraction of socially and clinically meaningful signals from patient-generated text, enabling downstream analysis of the patient voice at scale. The contributions of this work are as follows.
\begin{itemize}
\item We propose the PVminer Framework, a structured prediction formulation for extracting patient voice from patient-generated text that captures hierarchical labels, and evidence Spans. 


\item We propose a systematically designed prompt engineering approach and provide a benchmark of instruction-tuned large language models ranging from 1.5B to 70B parameters under zero-shot and few-shot settings, highlighting the limitations of prompt-based approaches for schema-constrained patient voice extraction.
\item We developed PVminerLLM, a set of supervised fine-tuned large language models with different sizes that achieve strong performance on the PVminer task. These LLMs  offer a practical and effective solution for extracting socially meaningful patient voice signals from unstructured text, regardless of their size.
\end{itemize}

\section{PVminer Task Formulation}\label{sec2}

\begin{figure*}[!h]
  \centering
    \includegraphics[width=\textwidth]{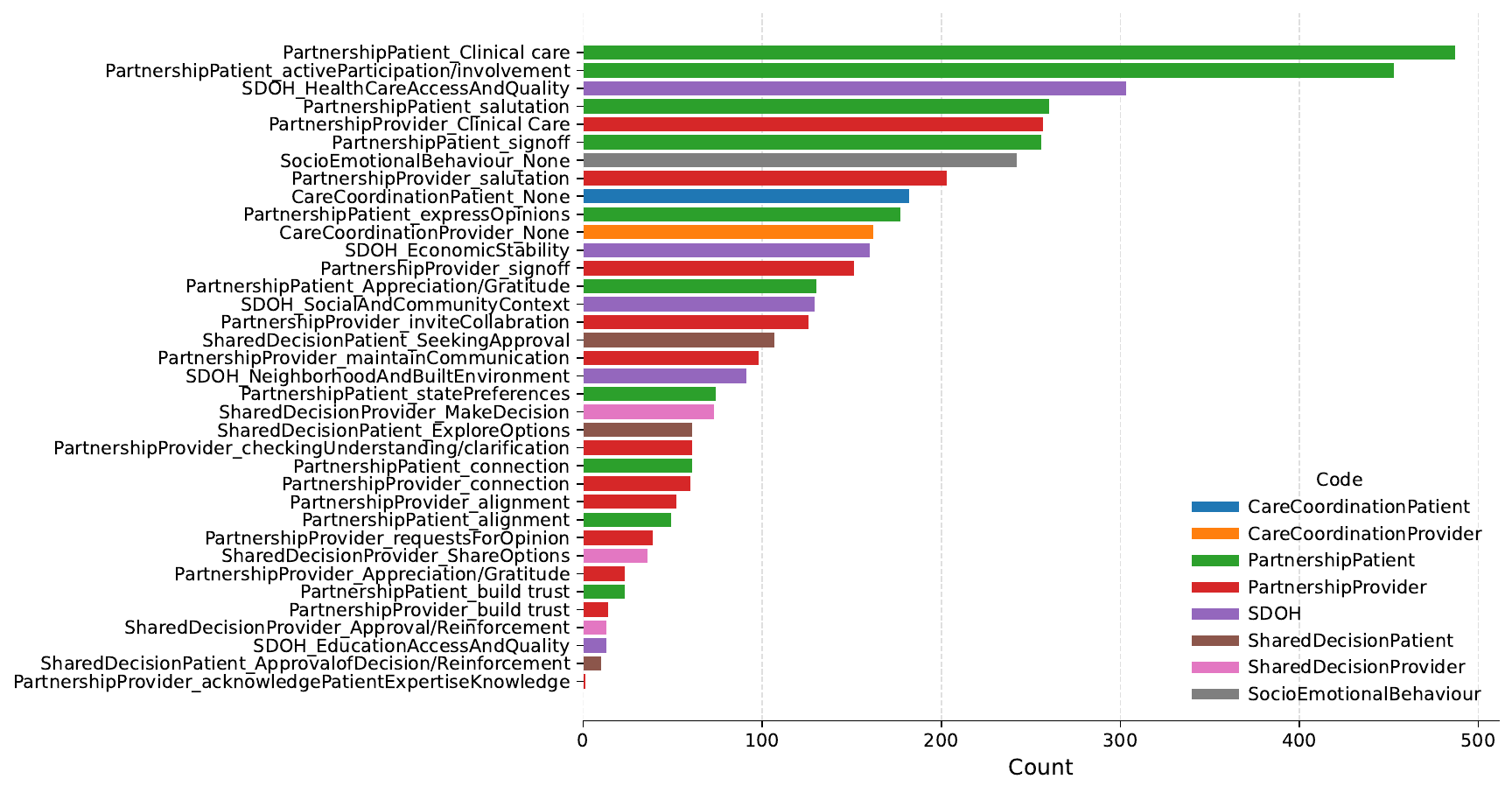}
  \caption{Distribution of Code–Sub-code pairs in the annotated dataset, colored by Code. Absent Sub-codes are labeled as None, which means this Code has no Sub-code in it. Zoom in for details.}
  \label{fig:combo_dist}
\end{figure*}

We define the PVminer task as a schema-constrained structured extraction problem over patient-generated text (i.e. portal messages). Given a single message, the task requires identifying all relevant patient voice expressions and representing each as a structured output consisting of a Code, a Sub-code, and a grounding text Span. This formulation is designed to support systematic evaluation of large language models under realistic constraints, where outputs must be both semantically correct and strictly schema-valid. Unlike conventional single-label or flat multi-label classification settings, PVminer allows multiple structured outputs to be extracted from a single message. Each output corresponds to a distinct expression that may reflect concerns, social context, or lived experience. The inclusion of Span grounding further requires models to localize evidence precisely within the input text, rather than relying on coarse semantic matching.

The labeling schema is organized hierarchically. Codes represent high-level semantic categories, while sub-codes capture more granular distinctions. A single Code may be associated with multiple Sub-codes, and certain Sub-codes may be shared across different Codes. This structure reflects the overlapping and non-exclusive nature of patient-generated language and introduces additional constraints that models must enforce during extraction. Definitions of all Codes and Sub-codes are provided in Appendix \ref{append:codebook}. The empirical distribution of each Code and Sub-code is shown in Fig.~\ref{fig:combo_dist}.

Formally, let $s$ denote a patient-generated message and let $d \in \{\texttt{Y}, \texttt{N}\}$ be a message-direction indicator, where $\texttt{Y}$ denotes provider to patient messages and $\texttt{N}$ denotes patient to provider messages. Let $\mathcal{C} = \{c_1, \dots, c_8\}$ denote the set of Codes and $\mathcal{U} = \{u_1, \dots, u_{26}\}$ denote the set of Sub-codes. A hierarchical constraint mapping $\mathcal{M}: \mathcal{C} \rightarrow 2^{\mathcal{U}}$ specifies the set of valid Sub-codes associated with each Code.

Given an input message, the extraction model $g_\phi$ produces a set of structured outputs:
\begin{equation*}
g_\phi(s, d) \;=\; \mathcal{E}(s)
\end{equation*}

\begin{equation*}
\mathcal{E}(s) \subseteq \mathcal{C} \times \mathcal{U} \times \mathcal{R}(s)
\end{equation*}

where $\mathcal{R}(s)$ denotes the set of all contiguous text Spans within $s$. Each extracted element $(c, u, r) \in \mathcal{E}(s)$ consists of a Code $c \in \mathcal{C}$, a Sub-code $u \in \mathcal{M}(c)$, and an evidence Span $r \in \mathcal{R}(s)$ that grounds the extracted label in the original text. 

A single message may yield one, or multiple such tuples, reflecting the variable density, overlap, and compositional nature of patient voice expressions in the text.



This task formulation serves as the foundation for benchmarking large language models under zero-shot and few-shot prompting, as well as for supervised fine-tuning to enforce schema adherence and improve structured extraction quality. In the following sections, we describe dataset construction and annotation, prompt-based extraction strategies, supervised fine-tuning for PVminer, and evaluation protocols tailored to structured, Span-grounded outputs.

\section{Datasets and Annotation}\label{sec3}

\begin{table}[!h]
\centering
\footnotesize
\caption{Demographic distribution across data sources. Percentages are calculated within each data source. }
\label{tab:demographics_transposed}
\renewcommand{\arraystretch}{1.1}
\setlength{\tabcolsep}{4.5pt}
\begin{tabular}{l|ccccc}
\hline
 & \textbf{YNHH} & \textbf{TXACC Woven} & \textbf{TXACC Bethesda} & \textbf{Survey} & \textbf{Total} \\
\hline
Sex (Male) & 81 (38\%) & 29 (48\%) & 71 (47\%) & 38 (26\%) & 219 (38\%) \\
Sex (Female) & 132 (62\%) & 32 (52\%) & 79 (53\%) & 109 (74\%) & 352 (62\%) \\
\hline
Race (White) & 152 (71\%) & 8 (13\%) & 20 (13\%) & 109 (74\%) & 289 (51\%) \\
Race (Black) & 35 (16\%) & 6 (10\%) & 15 (10\%) & 8 (5\%) & 64 (11\%) \\
Race (Asian) & 11 (5\%) & 4 (7\%) & 9 (6\%) & 19 (13\%) & 43 (8\%) \\
Race (Other) & 15 (7\%) & 43 (70\%) & 106 (71\%) & 11 (8\%) & 175 (31\%) \\
\hline
Ethnicity (Hispanic) & 30 (14\%) & 24 (39\%) & 60 (40\%) & 8 (5\%) & 122 (21\%) \\
\hline
\end{tabular}
\end{table}

Our study uses a corpus of patient-generated text (i.e. messages, and surveys) collected from multiple healthcare and research settings, including secure messaging from Yale New Haven Health, electronic messages from charitable clinics affiliated with the Texas Association for Charitable Clinics, and free-text patient survey responses from patient-centered outcomes research. Together, these sources capture diverse linguistic styles, care contexts, and social environments. The annotated corpus contains 1,137 messages, including 757 patient-authored and 380 provider-authored messages. In total, the dataset includes 46,038 word tokens, with an average message length of 40.5 words and a standard deviation of 32.8 words. Message length varies widely, ranging from brief clarifications with few words to long narrative descriptions, with the longest message containing 261 words, reflecting realistic properties of patient-generated text and posing meaningful challenges for structured extraction.

Although PVminer focuses on patient-generated text, provider-authored messages are intentionally included to preserve conversational context and enable models to distinguish intent and structure across message sources, reflecting real-world deployment conditions. Across all sources, the dataset includes message threads from 571 unique patients. Data from charitable clinics contribute linguistic, cultural, and socioeconomic diversity that complements messages from a large academic medical center, while survey responses broaden expression beyond clinical messaging systems. Using multiple data sources reduces institution-specific bias and improves generalization across heterogeneous settings. Training and testing splits are constructed using iterative stratification \cite{sechidis2011stratification} to preserve label coverage under the hierarchical, multi-label annotation schema. Demographic distributions across data sources are summarized in Table~\ref{tab:demographics_transposed}, highlighting the diversity of the patient population represented.

\subsection{PVminer vs. Clinical NLP Benchmarks}\label{benchmark:compare}

\begin{table*}[!h]
\centering
\footnotesize
\setlength{\tabcolsep}{4pt}
\renewcommand{\arraystretch}{1.1}

\caption{
Comparison between PVminer and representative biomedical NLP benchmarks.
$\checkmark$ indicates strong support, $\times$ indicates no support, and $\triangle$ indicates partial or limited support.
Benchmarks are compared along four properties relevant to patient voice extraction from patient-generated text. They are Relational/ Socio-Emotional, Bidirectional Interaction, Supports Multi-Label Coding and Tailored for Secure Messaging.
}

\label{tab:ben_compare_wide}

\begin{tabular}{p{1.6cm} p{1.6cm} p{2.0cm} p{1.6cm} c c c c}
\toprule
\textbf{Benchmark} &
\textbf{Domain} &
\textbf{Focus} &
\textbf{Data Source} &
\makecell{\textbf{Rel./}\\\textbf{Socio-}\\\textbf{Emo.}} &
\makecell{\textbf{Bidir.}\\\textbf{Interac.}} &
\makecell{\textbf{Multi-}\\\textbf{Label}} &
\makecell{\textbf{Secure}\\\textbf{Msg.}} \\
\midrule

CBLUE \cite{zhang2022cblue} &
Chinese biomedical &
NER, RE, classification &
Clinical text and dialogues &
$\times$ & $\times$ & $\times$ & $\times$ \\
\midrule
MedDG \cite{liu2022meddg}&
Chinese medical dialogue &
Diagnosis, symptom inquiry &
Simulated dialogues, semi-structured dialogues &
$\times$ & $\times$ & $\times$ & $\times$ \\
\midrule
ReMeDi \cite{yan2022remedi}&
English medical dialogue &
Medication intent, treatment reasoning &
Clinical conversations (non-SM) &
$\triangle$ & $\triangle$ & $\times$ & $\times$ \\
\midrule
MediTOD \cite{saley2024meditod}&
English history-taking &
Symptom elicitation, clinical reasoning &
Simulated structured dialogues &
$\times$ & $\times$ & $\times$ & $\times$ \\
\midrule
BLURB \cite{gu2021domain} &
Biomedical NLP &
NER, RE, QA, summarization &
Biomedical literature &
$\times$ & $\times$ & $\times$ & $\times$ \\
\midrule
\textbf{PVminer\-(Ours)} &
U.S. secure messaging  &
 relational behaviors, adherence cues &
De-identified SM from patient portals &
$\checkmark$ & $\checkmark$ & $\checkmark$ & $\checkmark$ \\
\bottomrule
\end{tabular}
\end{table*}

Table~\ref{tab:ben_compare_wide} compares PVminer with representative biomedical NLP benchmarks across four properties essential for patient voice extraction: relational and socio-emotional content, bidirectional interaction, multi-label annotation, and alignment with secure messaging data. Most existing benchmarks focus on narrowly scoped tasks using clinical notes, simulated dialogues, or biomedical literature, and therefore lack support for overlapping social signals and informal patient-authored language. Dialogue-oriented datasets introduce limited conversational structure but remain constrained to task-oriented or simulated settings, providing only partial coverage of relational and emotional expressions and typically assuming single-label annotations. In contrast, PVminer is explicitly designed for structured extraction from patient-generated secure messages, supporting multi-label, Span-grounded annotations and mixed-author message streams. These properties introduce challenges absent from prior benchmarks.

\subsection{Annotation}

We developed an annotation schema and a codebook to support structured extraction of patient voice from patient-generated text. The codebook captures socially and clinically meaningful expressions commonly found in secure messages and survey responses, while remaining compatible with schema-constrained modeling. All messages were annotated by domain experts in health communication and informatics using the eHOST platform \cite{yale2025annotation}, following an iterative protocol with regular collective reviewing, ambiguity handling and label definitions updating, which is shown in Fig.~\ref{fig:pipeline}. 

The final annotation schema represents patient voice using a two-level hierarchical structure. Each annotation consists of a Code capturing a high-level communicative or social function, a Sub-code specifying finer-grained intent or context, and a Span grounding the label in the original text. This design allows multiple overlapping annotations per message and supports fine-grained, Span-level evaluation. Figure~\ref{fig:combo_dist} shows that Code and Sub-code combinations are highly imbalanced, with a small number of frequent categories and a long tail of rare but semantically important cases. This reflects real-world patient-generated language and poses meaningful challenges for both prompt-based and supervised models. The schema comprises eight major Codes with associated Sub-codes; full definitions and examples are provided in Appendix~\ref{append:codebook}.

\section{Benchmark and Prompt Engineering}\label{sec4}


To provide a benchmark, we establish a baseline prompt within the PVminer framework for evaluating large language models as structured annotators. Unlike open-ended generation tasks, PVminer requires outputs to be parseable, label-consistent, and grounded in exact text Spans. Prompt design therefore focuses on reliability and constraint satisfaction rather than linguistic fluency. For a patient-generated message $s$ and a message-direction indicator $d$, models are prompted to produce schema-constrained outputs consisting of one or more Code, Sub-code, and Span tuples. Zero-shot is used to assess baseline performance, characterize task difficulty, and motivate subsequent supervised adaptation. The baseline prompt is described as Prompt 1 in Appendix ~\ref{append:prompt}.

Subsequently, to improve upon the baseline prompt, we designed a new prompt called the engineered prompt to elicit structured, multi-label annotations under strict schema constraints. In particular, the prompt explicitly specifies the output schema, enforces multi-label completeness, and requires that all extracted Spans be copied verbatim from the input message. Our prompt engineering task targets several dominant failure modes observed in zero-shot  structured extraction with the baseline prompt. These include format drift, where models produce non-parseable or verbose outputs; semantic confusion between closely related labels; errors arising from implicit assumptions about speaker role; and Span boundary noise, where extracted Spans are incomplete or hallucinated. To mitigate these issues, the prompt provides explicit label definitions, validity constraints on Code and Sub-code combinations, and direction-aware control signals that condition labeling decisions on message source (patient or provider). In addition, the prompt integrates structured reasoning guidance that scaffolds intermediate analytical steps, encouraging deliberate and schema-aligned labeling decisions. Rather than relying on one-shot label prediction, models are instructed to decompose the task into interpretation, label selection, and Span verification steps, while restricting the final output to a schema-valid format. This approach follows established prompt-pattern principles for complex decision tasks, emphasizing explicit structure, hard constraints, and internal verification. The engineered prompt is described as Prompt 2 in Appendix ~\ref{append:prompt}. While the engineered prompt improves schema validity and reduces common annotation errors, prompting alone remains insufficient for reliable extraction under the PVminer schema. In particular, rare labels, confusable Sub-codes, and token-critical Span boundaries continue to pose challenges, motivating supervised fine-tuning in subsequent section.

\section{PVminerLLM - Supervised Fine-Tuned LLMs}\label{sec5}

We developed PVminerLLM by applying supervised fine-tuning to specialize instruction-tuned language models for the PVminer task, with the goal of improving reliability in structured extraction under strict schema constraints. Each training instance consists of a patient-generated message $s$, a message-direction indicator $d \in \{\texttt{Y}, \texttt{N}\}$, and a gold annotation set $\mathcal{A}$ containing one or more structured tuples of the form $\{\text{Code}, \text{Sub-code}, \text{Span}\}$. During fine-tuning, the model learns to produce schema-valid outputs with accurate hierarchical label assignment and exact Span grounding. This training-based approach addresses reliability limitations observed in prompt-only inference. For training, each structured annotation set $\mathcal{A}$ is serialized into a JSON-formatted target string $a$. The conditioning query $q$ is formed by combining task instructions with the instance-specific message content and the message-direction indicator. The model is then trained to generate the structured completion $a$ conditioned on $q$, thereby learning to map patient-generated text to schema-conformant annotations.

Formally, the conditioning query $q$ is constructed as
\begin{equation*}
q = \mathcal{I} \,\Vert\, \texttt{\textbackslash n} \,\Vert\, s \,\Vert\, \texttt{\textbackslash n} \,\Vert\, d ,
\end{equation*}
where $\mathcal{I}$ denotes the task instruction template, $s$ denotes the patient-generated message text, $d$ denotes the message-direction indicator, and $\Vert$ represents string or token concatenation. The serialized target completion is defined as
\begin{equation*}
a = \mathrm{Serialize}(\mathcal{A}),
\end{equation*}
where $\mathrm{Serialize}(\cdot)$ maps the structured annotation set into a schema-valid JSON string.

Let $\mathbf{w} = [w_1, \ldots, w_L]$ denote the token sequence obtained by encoding the concatenation of the query $q$ and the target completion $a$. A binary mask $\mathbf{m} \in \{0,1\}^L$ is applied where $m_t = 1$ if token $w_t$ belongs to the serialized annotation $a$, and $m_t = 0$ otherwise. Tokens corresponding to task instructions and input context are excluded from optimization to ensure that learning focuses exclusively on structured output generation.

The supervised fine-tuning objective is defined as
\begin{equation*}
\mathcal{J}_{\mathrm{sup}}(\phi)
=
-\mathbb{E}_{(q,\mathcal{A}) \sim \mathcal{D}}
\left[
\frac{1}{\sum_{t=1}^{L} m_t}
\sum_{t=1}^{L}
m_t \log P_{\phi}(w_t \mid w_{<t})
\right],
\end{equation*}
where $P_{\phi}$ denotes the language model parameterized by $\phi$. This masked likelihood objective prevents the model from memorizing task instructions and instead allocates learning capacity to producing valid Code and Sub-code combinations and character-exact Span boundaries \cite{huerta-enochian-ko-2024-promptloss}.

We implement supervised fine-tuning using parameter-efficient adapters with QLoRA \cite{dettmers2023qlora}, applying low-rank updates to attention projection layers while keeping base model parameters fixed. This approach enables efficient adaptation across models of different sizes while maintaining computational feasibility. The resulting fine-tuned models, referred to as PVminerLLM, provide a practical and scalable solution for structured extraction of patient voice from patient-generated text.

\section{Experiments and Results}\label{sec6}

\subsection{Metrics}

We evaluate model performance using metrics tailored to the structured and multi-component nature of the PVminer task. Because each message may contain multiple overlapping labels and evidence Spans, evaluation is performed separately for Code prediction, Sub-code prediction, and Span extraction, with all metrics computed in a multi-label setting.

Code prediction is evaluated as a multi-label classification problem over the predefined set of Codes. Let $\hat{y}_i^{\mathrm{Code}}$ denote the set of Codes predicted for instance $i$, and let $y_i^{\mathrm{Code}}$ denote the corresponding gold standard set. Precision, recall, and F1-score are computed as
\begin{equation*}
\mathrm{precision}_{\mathrm{Code}} =
\frac{\sum_i \left| \hat{y}_i^{\mathrm{Code}} \cap y_i^{\mathrm{Code}} \right|}
     {\sum_i \left| \hat{y}_i^{\mathrm{Code}} \right|},
\end{equation*}

\begin{equation*}
\mathrm{recall}_{\mathrm{Code}} =
\frac{\sum_i \left| \hat{y}_i^{\mathrm{Code}} \cap y_i^{\mathrm{Code}} \right|}
     {\sum_i \left| y_i^{\mathrm{Code}} \right|},
\end{equation*}

\begin{equation*}
\mathrm{F1}_{\mathrm{Code}} =
\frac{2 \times \mathrm{precision}_{\mathrm{Code}} \times \mathrm{recall}_{\mathrm{Code}}}
     {\mathrm{precision}_{\mathrm{Code}} + \mathrm{recall}_{\mathrm{Code}}}.
\end{equation*}

Sub-code prediction is also evaluated as a multi-label classification task, where each message may be associated with multiple Sub-codes. Let $\hat{y}_i^{\mathrm{Sub}}$ and $y_i^{\mathrm{Sub}}$ denote the predicted and gold Sub-code sets for instance $i$, respectively. We compute
\begin{equation*}
\mathrm{precision}_{\mathrm{Sub}} =
\frac{\sum_i \left| \hat{y}_i^{\mathrm{Sub}} \cap y_i^{\mathrm{Sub}} \right|}
     {\sum_i \left| \hat{y}_i^{\mathrm{Sub}} \right|},
\end{equation*}

\begin{equation*}
\mathrm{recall}_{\mathrm{Sub}} =
\frac{\sum_i \left| \hat{y}_i^{\mathrm{Sub}} \cap y_i^{\mathrm{Sub}} \right|}
     {\sum_i \left| y_i^{\mathrm{Sub}} \right|},
\end{equation*}

\begin{equation*}
\mathrm{F1}_{\mathrm{Sub}} =
\frac{2 \times \mathrm{precision}_{\mathrm{Sub}} \times \mathrm{recall}_{\mathrm{Sub}}}
     {\mathrm{precision}_{\mathrm{Sub}} + \mathrm{recall}_{\mathrm{Sub}}}.
\end{equation*}

Evidence Span extraction is assessed using a relaxed token-level matching strategy designed to account for boundary ambiguity in natural language annotation. For each example $i$, let $\mathcal{S}^{(i)}_{\text{pred}}$ represent the set of predicted Spans and $\mathcal{S}^{(i)}_{\text{ref}}$ represent the set of reference Spans. A predicted Span $s_p \in \mathcal{S}^{(i)}_{\text{pred}}$ is considered a true positive if it aligns with at least one reference Span $s_r \in \mathcal{S}^{(i)}_{\text{ref}}$ according to any of the following conditions: the token-level Jaccard overlap between the two Spans is greater than or equal to 0.6.
Predicted Spans that fail to align with any reference Span are treated as false positives, while reference Spans without a corresponding prediction are treated as false negatives. Precision, recall, and F1-score for Span extraction are computed as


\begin{equation*}
\mathrm{precision}_{\mathrm{Span}} =
\frac{|\mathrm{TP}|}{|\mathrm{TP}| + |\mathrm{FP}|},
\end{equation*}

\begin{equation*}
\mathrm{recall}_{\mathrm{Span}} =
\frac{|\mathrm{TP}|}{|\mathrm{TP}| + |\mathrm{FN}|},
\end{equation*}

\begin{equation*}
\mathrm{F1}_{\mathrm{Span}} =
\frac{2 \times \mathrm{precision}_{\mathrm{Span}} \times \mathrm{recall}_{\mathrm{Span}}}
     {\mathrm{precision}_{\mathrm{Span}} + \mathrm{recall}_{\mathrm{Span}}}.
\end{equation*}

\subsection{Experimental Setting}

All experiments are conducted using the lm\_eval framework \cite{finben, eval-harness} with a vLLM backend. We evaluate instruction-tuned large language models across a range of sizes, including Llama-3.3-70B-Instruct, Llama-3.1-8B-Instruct, Llama-3.2-3B-Instruct \cite{llama3}, and Qwen2.5-1.5B-Instruct \cite{qwen2, qwen2.5}.

Prompt-based evaluation is performed in zero-shot and few-shot settings. Zero-shot experiments use a maximum context length of 8096 tokens, while few-shot experiments increase the context length to accommodate in-context exemplars. We use deterministic decoding for all evaluations with temperature set to zero and no sampling. Generation is constrained to schema-valid JSON outputs and terminates at a designated stop string (for example, \texttt{JSON\_END}), with a maximum of 1024 generated tokens per instance. Each model’s official chat template is applied at inference time, and a strict output contract is enforced to prevent extraneous text.

For supervised fine-tuning, we use parameter-efficient QLoRA adapters while keeping base model parameters frozen. Training uses a maximum input length of 8192 tokens with \texttt{bfloat16} precision, and the 70B model additionally applies 4-bit weight quantization. Gradient checkpointing and gradient accumulation are enabled to support long-context training. Optimization is performed with AdamW and a linear warmup schedule using the HuggingFace \texttt{Trainer}. All training runs are conducted on two H200 GPUs with distributed data parallelism.

\subsection{Baseline and Engineered Prompt Performance on the PVminer Benchmark}

\begin{table}[!h]
\centering
\caption{Comparison of F1 scores (\%) between Baseline  and Engineered prompts under zero-shot conditions. The baseline and engineered prompt variants are provided in Appendix~\ref{append:prompt}.}
\label{tab:prompt_engineering}
\small
\renewcommand{\arraystretch}{1.2}
\begin{tabular}{l l c c c}
\hline
\textbf{Model} & \textbf{Method} & \textbf{Code} & \textbf{Sub-code} & \textbf{Span} \\
\hline
\multirow{2}{*}{Llama-3.1-8B-Instruct}
 & Baseline          & 0.0   & 0.0    & 50.10 \\
 & Engineered-prompt & \textbf{47.09} & \textbf{20.84} & \textbf{54.15} \\
\hline
\multirow{2}{*}{Llama-3.3-70B-Instruct}
 & Baseline          & 57.71 & 27.24 & 47.20 \\
 & Engineered-prompt & \textbf{62.25} & \textbf{43.71} & \textbf{55.04} \\ 
\hline
\end{tabular}
\end{table}


Table~\ref{tab:prompt_engineering} reports F1 scores for Code, Sub-code, and Span extraction under a zero-shot setting. In both the baseline and engineered prompt configurations, no in-context examples (i.e., no 1-shot or few-shot demonstrations) were provided to the model. The two settings differ solely in the design of the instruction. The baseline prompt consists of a minimal task description with general guidance, whereas the engineered prompt instruction introduces structured output formatting, explicit decision logic, disambiguation rules, self-validation constraints, and performance-oriented guidance. This comparison isolates the effect of our proposed prompt engineering method (Section~\ref{sec4}) without altering the number of shots.

Across both models, the engineered prompt instruction consistently achieves higher F1 scores across all prediction targets. In particular, prompt engineering yields substantial gains in Code and Sub-code extraction for the 8B model and produces clear improvements across all tasks for the 70B model. These findings demonstrate that explicit structural guidance significantly enhances the reliability of multi-label, structured extraction in the PVminer benchmark, even under a strictly zero-shot regime. 

In addition to the overall F1 score improvements, from experiment observations, predictions on several commonly confused Codes and Sub-codes decreased as our proposed engineered prompt is introduced. For example, at the Code level, major swaps such as PartnershipPatient predicted as PartnershipProvider decrease from 18 to 5 cases, and the reverse direction decreases from 12 to 4 cases, reflecting clearer role distinction and more accurate label assignment overall. At the Sub-code level, large cross-confusions such as Salutation predicted as Signoff decrease from 24 cases to 7 cases, and Signoff predicted as Salutation decreases from 11 to 3 cases, reflecting better separation between opening and closing message components.

\subsection{Engineered Prompt  and Supervised Fine-tuning Performance}

\begin{table*}[!t]
\centering
\footnotesize
\setlength{\tabcolsep}{2pt}
\renewcommand{\arraystretch}{1.1}
\caption{Code-level performance of the engineered prompt and supervised fine-tuning (P/R/F1 in \%).}
\label{tab:code_all}
\begin{tabular}{lcccccccccccc}
\toprule
& \multicolumn{3}{c}{\textbf{Zero-shot}} 
& \multicolumn{3}{c}{\textbf{One-shot}} 
& \multicolumn{3}{c}{\textbf{Two-shot}} 
& \multicolumn{3}{c}{\textbf{SFT}} \\
\cmidrule(lr){2-4} \cmidrule(lr){5-7} \cmidrule(lr){8-10} \cmidrule(lr){11-13}
\textbf{Model} & P & R & F1 & P & R & F1 & P & R & F1 & P & R & F1 \\
\midrule
Llama-3.3-70B-Instruct 
& 69.98 & 56.06 & 62.25 
& 70.06 & 60.94 & 65.18 
& 70.88 & 59.86 & 64.90 
& 87.90 & 80.11 & \textbf{83.82} \\

Llama-3.1-8B-Instruct  
& 47.35 & 46.84 & 47.09 
& 60.83 & 47.74 & 53.50 
& 61.71 & 49.55 & 54.96 
& 85.04 & 78.12 & 81.43 \\

Llama-3.2-3B-Instruct  
& 42.83 & 34.54 & 38.24 
& 53.83 & 47.02 & 50.19 
& 54.91 & 47.56 & 50.97 
& 82.48 & 78.30 & 80.33 \\

Qwen2.5-1.5B-Instruct  
& 33.94 & 13.38 & 19.20 
& 46.36 & 21.88 & 29.73 
& 53.33 & 26.04 & 34.99 
& 83.19 & 71.61 & 76.97 \\

\bottomrule
\end{tabular}
\end{table*}


\begin{table*}[!t]
\centering
\footnotesize
\setlength{\tabcolsep}{2pt}
\renewcommand{\arraystretch}{1.1}
\caption{Sub-code-level performance of the engineered prompt and supervised fine-tuning (P/R/F1 in \%).}
\label{tab:subcode_all}
\begin{tabular}{lcccccccccccc}
\toprule
& \multicolumn{3}{c}{\textbf{Zero-shot}} 
& \multicolumn{3}{c}{\textbf{One-shot}} 
& \multicolumn{3}{c}{\textbf{Two-shot}} 
& \multicolumn{3}{c}{\textbf{SFT}} \\
\cmidrule(lr){2-4} \cmidrule(lr){5-7} \cmidrule(lr){8-10} \cmidrule(lr){11-13}
\textbf{Model} & P & R & F1 & P & R & F1 & P & R & F1 & P & R & F1 \\
\midrule
Llama-3.3-70B-Instruct 
& 50.68 & 38.42 & 43.71 
& 51.86 & 46.83 & 49.22 
& 53.69 & 48.90 & 51.18 
& 83.74 & 77.95 & \textbf{80.74} \\

Llama-3.1-8B-Instruct  
& 31.19 & 15.65 & 20.84 
& 48.47 & 28.72 & 36.07 
& 47.22 & 31.82 & 38.02 
& 79.19 & 76.33 & 77.73 \\

Llama-3.2-3B-Instruct  
& 15.77 & 11.38 & 13.22 
& 28.57 & 20.70 & 24.01 
& 32.19 & 24.19 & 27.62 
& 75.80 & 73.74 & 74.75 \\

Qwen2.5-1.5B-Instruct  
& 17.39 & 1.03 & 1.95 
& 34.31 & 9.06 & 14.33 
& 38.91 & 11.13 & 17.30 
& 77.86 & 66.88 & 71.96 \\

\bottomrule
\end{tabular}
\end{table*}


\begin{table*}[!t]
\centering
\footnotesize
\setlength{\tabcolsep}{2pt}
\renewcommand{\arraystretch}{1.1}
\caption{Span-level performance of the engineered prompt and supervised fine-tuning (P/R/F1 in \%).}
\label{tab:span_all}
\begin{tabular}{lcccccccccccc}
\toprule
& \multicolumn{3}{c}{\textbf{Zero-shot}} 
& \multicolumn{3}{c}{\textbf{One-shot}} 
& \multicolumn{3}{c}{\textbf{Two-shot}} 
& \multicolumn{3}{c}{\textbf{SFT}} \\
\cmidrule(lr){2-4} \cmidrule(lr){5-7} \cmidrule(lr){8-10} \cmidrule(lr){11-13}
\textbf{Model} & P & R & F1 & P & R & F1 & P & R & F1 & P & R & F1 \\
\midrule
Llama-3.3-70B-Instruct 
& 84.74 & 40.75 & 55.04 
& 86.69 & 53.82 & 66.41 
& 87.56 & 56.41 & 68.62 
& 88.02 & 86.07 & \textbf{87.03} \\

Llama-3.1-8B-Instruct  
& 72.30 & 43.28 & 54.15 
& 78.28 & 52.65 & 62.96 
& 74.10 & 62.45 & 67.78 
& 87.29 & 86.37 & 86.83 \\

Llama-3.2-3B-Instruct  
& 43.18 & 29.84 & 35.29 
& 58.01 & 41.31 & 48.25 
& 62.38 & 51.73 & 56.56 
& 85.29 & 84.34 & 84.81 \\

Qwen2.5-1.5B-Instruct  
& 70.38 & 12.45 & 21.16 
& 73.95 & 23.80 & 36.01 
& 72.31 & 43.16 & 54.05 
& 83.98 & 85.64 & 84.80 \\

\bottomrule
\end{tabular}
\end{table*}

Table~\ref{tab:code_all}-\ref{tab:span_all} report the engineered prompt zero-shot and few-shot performance on the PVminer benchmark across instruction-tuned models of varying sizes. For zero-shot, while larger models achieve moderate overall F1 scores, performance remains limited across all settings, particularly for Sub-code prediction and Span recall. Even for the largest model, recall for fine-grained labels is substantially lower than precision, reflecting difficulty in producing complete and schema-valid structured outputs under zero-shot prompting. These results indicate that zero-shot prompting alone is insufficient for reliable extraction on the PVminer task. One-shot and Two-shot provide incremental improvements. Besides the overall performance increasing, as the number of shot grows up, the difference between precision and recall is decreasing. But, one-shot and two-shot results still exhibit similar failure patterns. Recall is still significant lower than precision, even for largest model which motivates supervised fine-tuning.

Table~\ref{tab:code_all}-\ref{tab:span_all} also report performance after 10 epochs of supervised fine-tuning across models of varying sizes. Llama-3.3-70B-Instruct consistently outperforms other models achieving F1 scores of 83.82\% for Code classification, 80.74\% for Sub-code classification, and 87.03\% for Span extraction. Notably, Performance of the medium and smaller size models is comparable to the Llama-3.3-70B-Instruct model allowing for flexibility in using any of them as needed. Compared to zero- and few-shot prompting, supervised fine-tuning yields substantially higher precision, recall and F1-score across all tasks, with particularly strong gains for Code and Sub-code prediction. Compared to zero-shot learning, Llama-3.3-70B-Instruct achieves relative F1 improvements of 34.65\% (Code), 84.72\% (Sub-code), and 58.11\% (Span). Relative to 2-shot learning, the corresponding improvements are 29.15\%, 57.76\%, and 26.83\%, respectively.


\subsection {Patient Voice Domains Identified Using Two-Shot Engineered Prompt and PVminerLLM }

\begin{table*}[!h]
\centering
\small
\setlength{\tabcolsep}{3pt}
\renewcommand{\arraystretch}{1.1}
\caption{Two-shot and SFT performance of 70B model at the Code level (micro-averaged per class, in \%).}
\label{tab:code_twoshot_vs_sft}
\begin{tabular}{lcccccc}
\toprule
& \multicolumn{3}{c}{\textbf{Two-shot}} 
& \multicolumn{3}{c}{\textbf{SFT}} \\
\cmidrule(lr){2-4} \cmidrule(lr){5-7}
\textbf{Code} & P & R & F1 & P & R & F1 \\
\midrule
CareCoordinationPatient      
& 48.39 & 38.46 & 42.86 
& 74.36 & 74.36 & \textbf{74.36} \\

CareCoordinationProvider     
& 56.41 & 62.86 & 59.46 
& 78.79 & 74.29 & \textbf{76.47} \\

PartnershipPatient           
& 88.37 & 79.72 & 83.82 
& 91.73 & 85.31 & \textbf{88.41} \\

PartnershipProvider          
& 90.00 & 79.12 & 84.21 
& 91.86 & 86.81 & \textbf{89.27} \\

SDOH                         
& 87.50 & 45.90 & 60.22 
& 90.00 & 88.52 & \textbf{89.26} \\

SharedDecisionPatient        
& 31.82 & 40.00 & 35.44 
& 78.12 & 71.43 & \textbf{74.63} \\

SharedDecisionProvider       
& 33.33 & 34.62 & 33.96 
& 63.16 & 46.15 & \textbf{53.33} \\

SocioEmotionalBehaviour      
& 52.17 & 38.71 & 44.44 
& 81.97 & 80.65 & \textbf{81.30} \\

\bottomrule
\end{tabular}
\end{table*}


\begin{table*}[!h]
\centering
\small
\setlength{\tabcolsep}{3pt}
\renewcommand{\arraystretch}{1.05}
\caption{Two-shot and SFT performance of 70B model at the Sub-code level (micro-averaged per class, in \%).}
\label{tab:subcode_twoshot_vs_sft}
\begin{tabular}{lcccccc}
\toprule
& \multicolumn{3}{c}{\textbf{Two-shot}} 
& \multicolumn{3}{c}{\textbf{SFT}} \\
\cmidrule(lr){2-4} \cmidrule(lr){5-7}
\textbf{Sub-code} & P & R & F1 & P & R & F1 \\
\midrule
Appreciation/Gratitude & 46.94 & 74.19 & 57.50 & 96.77 & 96.77 & \textbf{96.77} \\
Approval/Reinforcement & 0.00 & 0.00 & \textbf{0.00} & 0.00 & 0.00 & \textbf{0.00} \\
ApprovalofDecision/Reinforcement & 0.00 & 0.00 & \textbf{0.00} & 0.00 & 0.00 & \textbf{0.00} \\
Clinical Care & 83.87 & 46.43 & 59.77 & 74.63 & 89.29 & \textbf{81.30} \\
EconomicStability & 81.82 & 23.08 & 36.00 & 86.84 & 84.62 & \textbf{85.71} \\
EducationAccessAndQuality & 100.00 & 50.00 & 66.67 & 100.00 & 100.00 & \textbf{100.00} \\
ExploreOptions & 23.33 & 58.33 & 33.33 & 58.33 & 58.33 & \textbf{58.33} \\
HealthCareAccessAndQuality & 55.17 & 25.00 & 34.41 & 74.60 & 73.44 & \textbf{74.02} \\
MakeDecision & 40.00 & 20.00 & 26.67 & 69.23 & 45.00 & \textbf{54.55} \\
NeighborhoodAndBuiltEnvironment & 57.14 & 36.36 & 44.44 & 75.00 & 54.55 & \textbf{63.16} \\
SeekingApproval & 22.22 & 16.67 & 19.05 & 80.95 & 70.83 & \textbf{75.56} \\
ShareOptions & 13.64 & 37.50 & 20.00 & 57.14 & 50.00 & \textbf{53.33} \\
SocialAndCommunityContext & 60.71 & 60.71 & 60.71 & 83.87 & 92.86 & \textbf{88.14} \\
acknowledgePatientExpertiseKnowledge & 0.00 & 0.00 & \textbf{0.00} & 0.00 & 0.00 & \textbf{0.00} \\
activeParticipation/involvement & 65.22 & 32.97 & 43.80 & 82.56 & 78.02 & \textbf{80.23} \\
alignment & 100.00 & 12.50 & 22.22 & 100.00 & 87.50 & \textbf{93.33} \\
build trust & 0.00 & 0.00 & 0.00 & 66.67 & 50.00 & \textbf{57.14} \\
checkingUnderstanding/clarification & 25.00 & 14.29 & 18.18 & 53.85 & 50.00 & \textbf{51.85} \\
connection & 22.58 & 26.92 & 24.56 & 81.48 & 84.62 & \textbf{83.02} \\
expressOpinions & 37.36 & 89.47 & 52.71 & 78.38 & 76.32 & \textbf{77.33} \\
inviteCollabration & 50.00 & 16.67 & 25.00 & 65.62 & 70.00 & \textbf{67.74} \\
maintainCommunication & 29.41 & 23.81 & 26.32 & 76.47 & 61.90 & \textbf{68.42} \\
requestsForOpinion & 33.33 & 54.55 & \textbf{41.38} & 60.00 & 27.27 & 37.50 \\
salutation & 90.82 & 91.75 & 91.28 & 100.00 & 98.97 & \textbf{99.48} \\
signoff & 68.09 & 75.29 & 71.51 & 97.62 & 96.47 & \textbf{97.04} \\
statePreferences & 26.92 & 50.00 & 35.00 & 71.43 & 71.43 & \textbf{71.43} \\
\bottomrule
\end{tabular}
\end{table*}

Within the engineered prompt, the two-shot setting yielded the highest performance, so we analyze the domains identified based on two-shot as representative performance and compare them with PVminerLLM results. From Tables \ref{tab:code_twoshot_vs_sft}–\ref{tab:subcode_twoshot_vs_sft},in the two-shot setting, performance differs substantially across patient voice domains and reflects their prevalence in the dataset (Fig. \ref{fig:combo_dist}). PartnershipPatient and PartnershipProvider are among the most prevalent domains and achieve the highest performance, with F1 scores of 83.82\% and 84.21\%, respectively. These domains capture frequent patterns of collaboration and involvement in patient-provider exchanges, which are often explicitly expressed in secure messages. Their strong performance suggests that two-shot prompting can reliably identify structurally clear patient voice signals. SDOH-related content is also prevalent, particularly sub-codes such as HealthCareAccessAndQuality and EconomicStability. For the SDOH code, the Two-shot engineered prompt achieves an F1 of 60.22\%, with high precision (87.50\%) but lower recall (45.90\%), indicating that socio-economic concerns are accurately detected when predicted but remain under-identified overall. CareCoordinationProvider, another relatively common domain, shows moderate performance (F1 = 59.46\%), whereas CareCoordinationPatient performs less consistently (F1 = 42.86\%). In contrast, SharedDecisionPatient and SharedDecisionProvider are less frequent in the corpus and exhibit substantially lower performance (F1 = 35.44\% and 33.96\%, respectively). At the Sub-code level, rare or nuanced categories such as alignment (F1 = 22.22\%), checkingUnderstanding/clarification (F1 = 18.18\%), and inviteCollabration (F1 = 25.00\%) show low recall and reduced overall performance, and several infrequent sub-codes collapse to 0.00\%. These findings indicate that two-shot prompting is more effective for prevalent and linguistically explicit patient voice domains but struggles to reliably identify low-frequency or context-dependent behaviors. Overall, domain-level variation in performance aligns with distributional imbalance, underscoring the need for task-specific model adaptation to improve detection of less common patient voice signals.

In contrast to the two-shot setting, PVminerLLM substantially improves identification across nearly all patient voice domains, particularly those that are both prevalent and clinically meaningful. At the code level, the most prevalent domains, PartnershipPatient and PartnershipProvider, reach F1 scores of 88.41\% and 89.27\%, respectively, improving from 83.82\% and 84.21\% under two-shot prompting. Similarly, the SDOH domain improves markedly from 60.22\% to 89.26\% F1, with recall increasing substantially, indicating that socio-economic concerns are no longer systematically under-identified. CareCoordinationPatient and CareCoordinationProvider also show meaningful gains (42.86\% to 74.36\% and 59.46\% to 76.47\%, respectively), reflecting stronger detection of logistical and treatment-related content. Notably, SharedDecisionPatient improves from 35.44\% to 74.63\% F1, and SharedDecisionProvider from 33.96\% to 53.33\%, suggesting that supervised fine-tuning better captures explicit decision-making behaviors that were previously difficult to identify with few-shot prompting. At the Sub-code level, improvements are especially significant for prevalent and clinically important signals. Clinical Care increases from 59.77\% to 81.30\%, EconomicStability from 36.00\% to 85.71\%, and SocialAndCommunityContext from 60.71\% to 88.14\%. High-frequency structural markers such as salutation and signoff remain strong (99.48\% and 97.04\% F1 under PVmineLLM), while several relational Sub-Codes that previously showed low recall under two-shot prompting demonstrate substantial gains, including activeParticipation/involvement (43.80\% to 80.23\%) and connection (24.56\% to 83.02\%). Although a small number of extremely rare Sub-Codes remain at or near zero performance, the overall pattern indicates that supervised fine-tuning reduces the performance gap between prevalent and less frequent patient voice domains.

\section{Discussion} \label{sec7}

\subsection{Key Insights from Prompting and Supervised Adaptation}

This study systematically examined the ability of instruction-tuned large language models to perform schema-constrained structured extraction on the PVminer task under prompt-based and supervised adaptation settings. Our experiments indicate that prompt-based inference alone cannot reliably achieve structured extraction under the PVminer constrained annotation schema, highlighting the need for additional mechanisms or fine-tuning. Even with carefully designed prompts, both zero-shot and few-shot settings show large gaps in recall, especially for Sub-code prediction and evidence Span extraction. Many errors come from outputs that do not fully follow the required structure, are incomplete, or mix up closely related labels. These findings suggest that while instruction-tuned models can often understand the general meaning of a message, they struggle to consistently produce fully structured outputs without task-specific supervision.

Supervised fine-tuning, on the other hand, leads to clear and consistent improvements across all parts of the task. The large gaps between precision and recall observed in zero-shot and few-shot performance (Table~\ref{tab:code_all}-\ref{tab:span_all}) are largely reduced after fine-tuning, indicating more complete and schema-consistent structured outputs. Span extraction achieves consistently high F1 scores across all models, and performance differences across model scales are reduced, supporting our claim that task-specific supervision enables reliable structured extraction on the PVminer task without reliance on extreme model size.
This indicates that the main challenge of PVminer is not just understanding the text itself, but also learning how to reliably translate patient-generated language into structured, Span-based annotations under strict formatting rules. Compared to few-shot prompting which can only identify highly frequent and structurally explicit domains, supervised fine-tuning enables more consistent and balanced detection across both prevalent and nuanced patient voice behaviors. This finding is particularly important for domains such as SDOH and shared decision-making, which carry significant implications for understanding patient needs and support.

Furthermore, the effect of model size becomes much smaller after fine-tuning. While larger models still perform slightly better, smaller models reach comparable performance once adapted to the task. This suggests that successful patient voice extraction depends more on task alignment and high-quality supervision than on model size alone. As a result, accurate extraction can be achieved with smaller models of the PVminerLLM's suite of supervised fine-tuned LLMs, making large-scale and practical deployment more feasible in healthcare settings.

\subsection{Clinical and Social Implications}

The ability to accurately extract the patient voice domains from patient-generated text has important implications for patient-centered care and health equity. Many signals that matter for clinical outcomes are social rather than purely medical \cite{intro1, intro2, intro3, intro4, intro5, intro6, intro7}. Patients often describe emotional stress, difficulty paying for care, transportation problems, caregiving responsibilities, or uncertainty about treatment decisions in their own words. These experiences shape whether patients can follow treatment plans, stay engaged with care teams, and feel supported, yet they are rarely captured in structured clinical records and are therefore easy to overlook in routine care. By focusing on unstructured patient-generated text, our proposed PVminerLLM that defines and captures Codes such as Social And Community Context, Neighborhood and Built Environment, Socioeconomic Status helps make these social and lived experiences visible at scale. Instead of relying on manual chart review or small qualitative studies, health systems can systematically extract and summarize patient voice signals across large populations. This makes it possible to identify patterns that are difficult to see otherwise, such as common barriers to adherence, frequent sources of confusion or distress, or groups of patients who may need additional support beyond standard clinical care.

In clinical practice, these capabilities can support more informed and responsive care. Structured patient voice information can help care teams recognize when patients are struggling with social or emotional challenges, even if those issues are not explicitly raised during visits. It can also guide targeted interventions, such as referrals to social services, adjustments to care plans, or additional follow-up for patients at risk of disengagement. By better aligning care with patients’ real-world circumstances, health systems can improve both effectiveness and patient experience.

Importantly, our results show that scalable patient voice extraction does not require extremely large models or highly specialized data. This makes the approach more accessible to a wide range of healthcare settings, including community clinics and resource-constrained systems. From a research perspective, PVminerLLM provides essential infrastructure for incorporating patient voice into patient-centered outcomes research. By enabling large-scale analysis of social context and lived experience alongside clinical information, this work supports more complete assessments of care quality, equity, and effectiveness, and helps ensure that patients’ voices are meaningfully represented in data-driven healthcare decisions.

\subsection{Future Work}

The proposed PVminerLLM  framework demonstrates strong capability in extracting and structuring patient voice content, effectively mapping patient-generated language to clinically and socially meaningful domains. However, we plan to extend this work to enhance its practical and clinical deployment and reduce its complexity. First, the current prompt design is necessarily large and complex due to the hierarchical schema, strict output constraints, and need for disambiguation rules. A promising next step is the use of multi-agent or modular inference frameworks, where distinct agents handle complementary subtasks such as semantic interpretation, label selection, and Span verification. Decomposing the task in this way may reduce prompt complexity while improving robustness and interpretability \cite{multi1, multi2, multi3, multi4, multi5, multi6}. Second, while supervised fine-tuning yields strong performance, post-SFT alignment remains underexplored for structured extraction tasks like PVminer. Preference-based or constraint-aware alignment methods tailored to token-critical outputs could further improve reliability, particularly for rare Sub-codes and boundary-sensitive Spans \cite{ouyang2022training, bai2022constitutional, gheshlaghiAzar2024psiPO, meng2024simpo, xiao2024caldpo, pal2024smaug}. We aim to further strengthen the reliability, scalability, robustness and clinical utility of patient voice extraction systems, moving closer to practical deployment in real-world healthcare settings.

\section{Conclusion}\label{sec7}

In this work, we introduced the PVminer task and benchmark for structured extraction of patient voice from patient-generated text. By formalizing patient voice annotation as a schema-constrained task with hierarchical labels and Span grounding, we provided a rigorous testbed for evaluating large language models under realistic clinical constraints. Our benchmark results show that prompt-based inference alone is insufficient for reliable extraction, while supervised fine-tuning yields substantial and consistent improvements across models of varying sizes.

The proposed PVminerLLM models demonstrate that accurate and scalable patient voice extraction can be achieved without reliance on extreme model scale, making deployment more feasible in real-world healthcare settings. By enabling systematic measurement of social and experiential signals embedded in patient-generated text, PVminer supports more holistic patient-centered research and lays the groundwork for integrating the patient voice into data-driven clinical and health services applications.

\backmatter





\bmhead{Conflicts of Interest}

No competing interest is declared.



\bmhead{Authors Contribution Statement}

S.F. conceptualized the study, designed the methodology and data analysis plan, and led the writing and revision of the manuscript. L.M. conducted the experiments and the analysis, contributed to the interpretation of the results, and co-wrote the manuscript. G.P., S.T., and A.K. assisted with conducting the experiments. A.H., S.L., and A.R. provided domain expertise in patient care and supported the accuracy and integrity of the content. All authors reviewed and approved the final manuscript.

\bmhead{Data Availability Statement}

The data analyzed in this study consist of de-identified patient–provider secure messages and associated annotations derived from clinical systems. Due to privacy, ethical, and institutional restrictions, these data are not publicly available. Access to the data may be considered upon reasonable request to the corresponding author and with appropriate institutional review board (IRB) approval and data use agreements in place.

\bmhead{Funding}

This work was supported by the Patient-Centered Outcomes Research Institute (PCORI) under Award No. ME-2023C2-31367 (to S.F.).










\bibliography{sn-bibliography}
\newpage
\begin{appendices}

\section{Instruction Prompt Design}\label{append:prompt}

\newtcolorbox{promptboxone}[1][]{
  enhanced,
  breakable,
  colbacktitle=green!70!black,
  colframe=green!60!black,
  colback=green!4!white,
  coltitle=white,
  title={Prompt 1 - Baseline Prompt},
  boxrule=0.9pt,
  arc=2mm,
  left=6pt,right=6pt,top=6pt,bottom=6pt,
  #1
}

\begin{promptboxone}
\begin{Verbatim}[
  breaklines=true,
  breakanywhere=true,
  breaksymbolleft={},
  breaksymbolright={},
  xleftmargin=0pt,
  formatcom=\ttfamily\footnotesize\color{black},
  commandchars=\\\{\}
]
You are a patient-centered communication analyst tasked with identifying and classifying how patients and clinicians incorporate patient-centered communication (PCC) elements in secure messaging.

Your goal is to extract multiple Code-Sub-code pairs from the current sentence and identify specific spans corresponding to each pair. This task requires careful, step-by-step reasoning to ensure accurate multi-label classification, with additional consideration of contextual information from surrounding sentences.

## Follow these steps systematically and step-by-step Instructions:
1. Understand the Input Sentence:
1.1 Analyze the message to establish the full context.
1.2 Note:
1.3 Carefully read and analyze every word in the message to determine its context and identify all relevant communication elements.
2. Identify Relevant Codes:
2.1 Match parts of the message to one or more Codes based on the intent and content described in the definitions below.
2.3 Acknowledge that a message may involve multiple Codes.
3. Determine Sub-codes for Each Code:
3.1 For each identified Code, assign the appropriate Sub-code(s) that further specify the meaning.
3.2 Use definitions of Sub-codes to ensure accuracy and consistency.
3.4 Important: Ensure that the Sub-code you select belongs to the Sub-code list under the identified Code. If it doesn't, reconsider whether the Code or Sub-code selection is correct.
4. Pair Codes with Sub-codes:
4.1 Form unique Code-Sub-code pairs for the message. These pairs should fully describe the meaning of the message.
4.2 If multiple Codes exist in the same message, their Sub-codes will differ.
5. Highlight Evidence for Each Pair:
5.1 Extract minimal, specific spans of text from the message that support each identified Code-Sub-code pair.
5.2 Note: The extracted minimum span should be a core phrase in the message instead of the entire sentence.

The following content provides definitions for Codes and Sub-Codes.

## Code and Definitions:

(\textit{Full list omitted here for brevity}).

FORMAT (Code WITH Sub-codes):
CODE_NAME: <one-sentence operational definition>.
|- SUBCODE_1: <one-sentence operational definition>.
|- SUBCODE_2: <one-sentence operational definition>.
|- SUBCODE_K: <one-sentence operational definition>.

FORMAT (Code WITHOUT Sub-codes):
CODE_NAME: <one-sentence operational definition>.
|- None: No sub-codes are defined for this Code.

Ensure your reasoning is step-by-step to capture all relevant Code-Sub-code pairs and their corresponding spans accurately. Remember, the Sub-code must belong to the list of Sub-codes under the identified Code.

IMPORTANT: Output your final result without any explanation and reasoning, you must output the JSON format like {"results": [{"Code": "<Identified Code>_1", "Sub-code": "<Identified Sub-code>_1", "Span": "<Extracted span>_1"},...,{"Code": "<Identified Code>_n", "Sub-code": "<Identified Sub-code>_n", "Span": "<Extracted span>_n"}]}
\end{Verbatim}
\end{promptboxone}

\newtcolorbox{promptbox}[1][]{
  enhanced,
  breakable,
  colbacktitle=purple!70!black,
  colframe=purple!60!black,
  colback=purple!4!white,
  coltitle=white,
  title={Prompt 2 - Engineered Prompt},
  boxrule=0.9pt,
  arc=2mm,
  left=6pt,right=6pt,top=6pt,bottom=6pt,
  #1
}

\newtcolorbox{promptboxwotitle}{
  enhanced,
  breakable,
  colframe=yellow!60!black,
  colback=yellow!4!white,
  boxrule=0.9pt,
  arc=2mm,
  left=6pt,right=6pt,top=6pt,bottom=6pt,
}

\begin{promptbox}[colbacktitle=yellow!70!black,
  colframe=yellow!60!black,
  colback=yellow!4!white]
\ttfamily\footnotesize
\vspace{0.4em}
\begin{legendbox}
\LegendItem{techXML}{XML Structuring}\\
\LegendItem{techCOT}{Chain-of-Thought (4-step reasoning)}\\
\LegendItem{techVAL}{Self-Validation (quality gate)}\\
\LegendItem{techLOG}{Decision Logic (direction-aware)}\\
\LegendItem{techDIS}{Disambiguation Rules}\\
\LegendItem{techPERF}{Performance Targets}
\end{legendbox}

\begin{alltt}
\XML{<role>}
Expert patient-centered communication analyst with \PERF{>95\%} accuracy in medical message \\ multi-label classification.
\XML{</role>}

\XML{<performance_target>}
\PERF{CRITICAL REQUIREMENTS:}
\PERF{- Code Accuracy: >95\%}
\PERF{- Sub-code Accuracy: >95\%}
\PERF{- Span Accuracy: >98\% (character-perfect)}

\PERF{Every annotation must be defensible and verification-validated.}
\XML{</performance_target>}

\XML{<task>}
Extract Code, Sub-code, and Span triples from medical secure messages.

INPUT:
- Context (the message text)
- Message Direction (\LOGIC{TO\_PAT\_YN})

OUTPUT:
- JSON list of \{Code, Sub-code, Span\} objects

CONSTRAINT:
- MULTI-LABEL task (one message may contain multiple valid triples)
\XML{</task>}

\XML{<message_direction>}
\LOGIC{CRITICAL: Message direction determines Code selection.}

\LOGIC{- TO\_PAT\_YN = "Y": Provider speaking TO patient}
\LOGIC{- TO\_PAT\_YN = "N": Patient speaking TO provider}

\LOGIC{USE CASES:}
\LOGIC{- Provider to patient: Use PartnershipProvider, SharedDecisionProvider,}
\LOGIC{  CareCoordinationProvider when applicable}
\LOGIC{- Patient to provider: Use PartnershipPatient, SharedDecisionPatient,}
\LOGIC{  CareCoordinationPatient when applicable}
\LOGIC{- SDOH and SocioEmotionalBehaviour apply regardless of direction}
\XML{</message_direction>}

\XML{<critical_rules>}
\VAL{RULE VIOLATIONS RESULT IN ANNOTATION FAILURE}

\VAL{1. Span Source:}
\VAL{   - Extract Spans ONLY from the provided message text}
\VAL{   - Context is for understanding only}
\VAL{   - Never invent, paraphrase, or infer Spans}

\VAL{2. Code and Sub-code Validity:}
\VAL{   - Every Sub-code MUST be valid for its Code}
\VAL{   - If a pairing is illogical or invalid, loop back and re-select}
\end{alltt}
\end{promptbox}
\begin{promptboxwotitle}
\ttfamily\footnotesize
\begin{alltt}
\VAL{3. Span Exactness:}
\VAL{   - Copy EXACT text from the message}
\VAL{   - Preserve punctuation, capitalization, and spacing}
\VAL{   - No paraphrasing}

\VAL{4. Multi-label Requirement:}
\VAL{   - Identify ALL relevant Code and Sub-code pairs in the message}
\XML{</critical_rules>}

\XML{<reasoning_process>}
\COT{Follow this 4-step verification process:}

\COT{STEP 1: CONTEXT AND DIRECTION ANALYSIS}
\COT{- Read the full message carefully}
\COT{- Determine message direction using TO\_PAT\_YN}
\COT{- Understand speaker intent and conversational goal}

\COT{STEP 2: PHRASE DECOMPOSITION AND CODE MATCHING}
\COT{- Break the message into semantic units (phrases or clauses)}
\COT{- For each phrase, identify intent:}
\COT{  * SDOH}
\COT{  * PartnershipProvider or PartnershipPatient}
\COT{  * SharedDecisionProvider or SharedDecisionPatient}
\COT{  * CareCoordinationProvider or CareCoordinationPatient}
\COT{  * SocioEmotionalBehaviour}
\COT{- Use TO\_PAT\_YN to select Provider vs Patient variants}
\COT{- Match each phrase to the correct Code definition}
\COT{- Verify Code and Sub-code pairing is logical and valid}

\COT{STEP 3: SPAN EXTRACTION AND VERIFICATION}
\COT{- Extract the MINIMUM complete supporting phrase}
\COT{- Spans must come ONLY from the message text}
\COT{- Verify character-level exactness}
\COT{- If the Span does not exist exactly, reject the annotation}

\COT{STEP 4: CROSS-VALIDATION (MOST IMPORTANT)}
\COT{Verification priority:}
\COT{1. Best semantic match confirmed (if not, loop back to Step 2)}
\COT{2. Sub-code valid for Code}
\COT{3. Span is exact and present in message}
\COT{4. All relevant phrases analyzed}
\COT{5. Disambiguation rules applied correctly}
\COT{6. High-confidence annotation defensible to experts}
\XML{</reasoning_process>}

\XML{<codes_definitions>}
The following are authoritative ground-truth definitions. Names must match exactly.
(\textit{Full list omitted here for brevity}).

FORMAT (Code WITH Sub-codes):
CODE_NAME: <one-sentence operational definition>.
|- SUBCODE_1: <one-sentence operational definition>.
|- SUBCODE_2: <one-sentence operational definition>.
|- SUBCODE_K: <one-sentence operational definition>.

FORMAT (Code WITHOUT Sub-codes):
CODE_NAME: <one-sentence operational definition>.
|- None: No sub-codes are defined for this Code.
\XML{</codes_definitions>}

\XML{<disambiguation_rules>}
\DIS{Apply systematically to resolve ambiguity.}

\DIS{Salutation vs Signoff:}
\DIS{- Opening greetings indicate salutation}
\DIS{- Closing phrases indicate signoff}
\DIS{- Position determines classification}
\end{alltt}
\end{promptboxwotitle}
\begin{promptboxwotitle}
\ttfamily\footnotesize
\begin{alltt}
\DIS{Appreciation/Gratitude vs Signoff:}
\DIS{- Simple closing thanks indicates signoff}
\DIS{- Specific appreciation indicates Appreciation/Gratitude}

\DIS{Provider vs Patient Codes:}
\DIS{- Use TO\_PAT\_YN strictly}
\DIS{- TO\_PAT\_YN = "Y" -> Provider codes}
\DIS{- TO\_PAT\_YN = "N" -> Patient codes}

\DIS{SharedDecision Codes:}
\DIS{- Use TO\_PAT\_YN to select Provider vs Patient variants}

\DIS{SDOH Sub-code Selection:}
\DIS{- EconomicStability: finances, income, food, housing}
\DIS{- EducationAccessAndQuality: education, literacy}
\DIS{- HealthCareAccessAndQuality: access to care, insurance, physical activity}
\DIS{- NeighborhoodAndBuiltEnvironment: housing, transportation, environment}
\DIS{- SocialAndCommunityContext: social support, isolation, discrimination}

\DIS{CareCoordination vs maintainCommunication:}
\DIS{- maintainCommunication: future updates only}
\DIS{- CareCoordination: concrete coordination with other providers}

\DIS{requestsForOpinion vs inviteCollaboration:}
\DIS{- requestsForOpinion: asks patient views}
\DIS{- inviteCollaboration: invites joint participation}

\DIS{SocioEmotionalBehaviour:}
\DIS{- Emotional support, reassurance, empathy, politeness}
\DIS{- Only Sub-code "None" is valid}
\XML{</disambiguation_rules>}

\XML{<output_format>}
Return JSON with a "results" array:

\{
  "results": [
    \{
      "Code": "exact Code name",
      "Sub-code": "exact Sub-code name",
      "Span": "EXACT text from message"
    \}
  ]
\}

If no annotations apply:
\{"results": []\}
\XML{</output_format>}

\XML{<quality_gate>}
\VAL{MANDATORY verification before submission:}

\VAL{1. JSON is parseable}
\VAL{2. All Sub-codes valid for Codes}
\VAL{3. All Spans are exact and present in message}
\VAL{4. Best semantic match verified}
\VAL{5. All disambiguation rules applied}
\VAL{6. High confidence suitable for expert review}

\VAL{Accuracy is paramount. Quality over speed.}
\XML{</quality_gate>}

INPUT:
\LOGIC{TO\_PAT\_YN: N (Patient speaking to provider)}
\end{alltt}
\end{promptboxwotitle}
\begin{promptboxwotitle}
\ttfamily\footnotesize
\begin{alltt}
Context:
Dr. Person1 I need my prescription sent to the pharmacy for my flecainide acetate
100 mg tablets twice a day the pharmacist has try requesting it no success and I
don't have any pills. Person2
\end{alltt}
\end{promptboxwotitle}


\section{Detailed CodeBook}\label{append:codebook}

The annotation schema consists of eight major Codes, each representing a distinct communicative or social construct within the patient voice. Each major code has a corresponding set of Sub-codes that capture more specific communicative intents. Below are concise summaries of these key categories:

\vspace{0.5em}
\noindent\textbf{1. Social Determinants of Health (SDOH).}  
Refers to the process of sharing and seeking knowledge about the social, economic, and environmental factors that significantly influence an individual’s health and well-being. These include economic stability, access and quality to education, access and quality of health care, neighborhood and built environment, and social and community context. 
\textit{Subcodes:}
\begin{itemize}
    \item \textbf{Economic Stability:} Financial security and resources to afford healthcare, housing, food, and other necessities. 
    \item \textbf{Education Access and Quality:} Provide information on the availability and effectiveness of educational opportunities, the quality of education, including resources, teaching standards, and curriculum.
    \item \textbf{Access and quality to healthcare:} The ability to obtain the necessary health services and the effectiveness and standard of care that patients receive.
    \item \textbf{Neighborhood and Built Environment:} Physical and social surroundings such as housing, transportation, safety, and environmental quality. 
    \item \textbf{Social and Community Context:} Social and community context refers to relationships, interactions, and conditions within the environments where people live, work, and interact, and it significantly influences health outcomes.
\end{itemize}

\noindent\textbf{2. Partnership from the Patient Side.}  
Patient partnership involves establishing and strengthening the alliance between patients and healthcare providers through active participation, open communication, and mutual respect.  
\textit{Subcodes:}
\begin{itemize}
    \item \textbf{Active Participation/Involvement:} Patient is active in providing information to aid diagnosis and problem solving,  priorities for treatment or management, asking questions and/or contributing to the identification of management approaches. 
    \item \textbf{Express Opinions:} Patients actively share their thoughts, concerns about their care and treatment and provide feedback on their experiences with healthcare services. 
    \item \textbf{Signoff:} Courteous closure that marks the completion of a message. 
    \item \textbf{State Preferences:} Individual values, desires, and priorities of the patient regarding their healthcare. 
    \item \textbf{Alignment:} Refers to establishment of a meaningful, trust-based relationship between the patient and healthcare provider. 
    \item \textbf{Appreciation/Gratitude:} appreciation and gratitude that are expressed when patients acknowledge the care and support they receive. 
    \item \textbf{Connection:} Information that is not directly related to the medical issue being discussed, strengthening the relationship among between the patient and provider.  
    \item \textbf{Salutation:} Greeting or addressing the provider by name or title. 
    \item \textbf{Clinical Care:} Refers to the patient’s experiences, perceptions, and self-reported expressions concerning symptoms, diagnoses, treatments, or medical procedures they receive or seek.
    \item \textbf{Build Trust:} Fostering a sense of confidence and reliability, wherein the patient feels assured that the provider is acting in their best interest. 
\end{itemize}

\noindent\textbf{3. Partnership from the Provider Side.}  
Refers to fostering a collaborative and equitable relationship between healthcare providers and patients, involving the equalization of status and ensuring that patients feel valued and empowered. 
\textit{Subcodes:}
\begin{itemize}
    \item \textbf{Invite Collaboration:} Inviting patients to participate in decisions related to their condition through informed consent, treatment planning, and self-management. 
    \item \textbf{Requests for Opinion:} actively seeking patients’ perspectives and preferences on treatment options, care plans, and health-related decisions. 
    \item \textbf{Checking Understanding/Clarification:} Confirming that the patient fully understands the information being communicated, including key concepts related to their condition, treatment options, costs, and care plan. 
    \item \textbf{Appreciation/Gratitude:} Refers to the expression of appreciation and gratitude by healthcare providers when acknowledging patients’ engagement, cooperation, and participation in their care.  
    \item \textbf{Signoff:} Courteous ending signaling completion of communication. 
    \item \textbf{Acknowledge Patient Expertise/Knowledge:} Recognizing and valuing the insights patients gain from their lived experiences with their health conditions. 
    \item \textbf{Maintain Communication:} It involves keeping patients informed and engaged throughout the care process by clearly communicating that additional information or updates will be provided at a later time, ensuring patients are aware of what to expect. 
    \item \textbf{Alignment:} Confirming or seeking confirmation that the patient and provider share a mutual understanding and perspective on a given issue. 
    \item \textbf{Connection:} Comments or information unrelated to the medical issue that serve to strengthen the patient–provider relationship and foster rapport.
    \item \textbf{Salutation:} Refers to the act of greeting or respectfully addressing the patient. 
    \item \textbf{Clinical Care:} Refers to the provider’s planning, coordination, and delivery of diagnoses, treatments, and medical procedures tailored to the patient’s clinical needs. 
    \item \textbf{Build Trust:} Involves fostering a relationship with the patient characterized by mutual respect, transparency, and reliable communication, which is essential for effective healthcare delivery. 
\end{itemize}

\noindent\textbf{4. Shared Decision-Making from the Patient Side.}  
Involves patients actively engaging in treatment discussions, asking questions, and providing informed approval of decisions that align with their values and goals, while fostering confidence in the chosen course of action.  
\textit{Subcodes:}
\begin{itemize}
    \item \textbf{Explore Options:} Indicators of a patient’s interest in learning about all options include asking detailed questions, expressing curiosity about alternatives, seeking clarification, discussing how options align with personal goals, and requesting additional resources for further information. 
    \item \textbf{Seeking Approval:} Refers to the patient asking for or seeking permission from the healthcare provider.
    \item \textbf{Approval of Decision:} Affirmation refers to the patient agreeing with a chosen plan or decision after understanding the available options and their implications. 
\end{itemize}

\noindent\textbf{5. Shared Decision-Making from the Provider Side.}  
Refers to statements or questions that create an opportunity for a two-way exchange of information regarding treatment options, evaluation of decisions, or consideration of alternatives.  
\textit{Subcodes:}
\begin{itemize}
    \item \textbf{Share Options:} Refers to the process of identifying, presenting, and deliberating the available options for a specific decision or problem, encompassing a discussion of their associated risks, benefits, and potential outcomes. 
    \item \textbf{Summarize and Confirm Understanding:} Involves the provider or decision-maker recapping the discussion using clear and concise language, summarizing options, risks, benefits, and other relevant details, while acknowledging the patient’s expressed values and concerns, and verifying the patient’s comprehension of the information. 
    \item \textbf{Make Decision:} Refers to the collaborative process in which the provider and patient jointly select a course of action after considering all available options, associated risks and benefits, and the patient’s personal preferences.
    \item \textbf{Approval of Decision/Reinforcement:} Refers to reinforcing a decision by providing positive feedback or validating the patient’s involvement in the decision-making process. 
\end{itemize}

\noindent\textbf{6. Socioemotional Behavior.}  
Refers to responding to patients’ emotional cues by validating their feelings, offering empathy, reassurance, and encouragement, acknowledging errors transparently, and providing positive reinforcement, reward, or approval. No Sub-codes are defined.

\noindent\textbf{7. Care Coordination Provider.}  
Refers to the deliberate organization and integration of healthcare services by the provider to ensure that patient care is delivered effectively, efficiently, and in a timely manner. No Sub-codes are defined.

\noindent\textbf{8. Care Coordination Patient.}  
Represents the patient’s active engagement in communicating and coordinating care among various healthcare providers and services to ensure continuity and effectiveness. No Sub-codes are defined.




\end{appendices}


\end{document}